\documentclass[11pt,a4paper]{article}
\usepackage[T1]{fontenc}
\usepackage{graphicx}
\usepackage{tikz}
\usepackage{pgfplots}  
\pgfplotsset{compat=1.18} 
\usepgfplotslibrary{statistics}
\usepackage{url}
\usepackage{hyperref}
\usepackage[nameinlink]{cleveref}
\usepackage{subcaption}
\usepackage{authblk}
\usepackage{hyperref}

\usepgfplotslibrary{statistics}

\newcommand{\mreal}{REAL}
\newcommand{\msynth}{SYNTH}
\newcommand{\mmix}{MIX}
\newcommand{\msampler}{SAMPLER}
\newcommand{\mcare}{CARE}

\begin{document}

\title{Synthetic Data for Robust Runway Detection}

\author[1,2]{\href{https://orcid.org/0009-0001-8659-6006}{Estelle Chigot}\thanks{Corresponding author}}
\author[1]{Dennis G. Wilson}
\author[2]{Meriem Ghrib}
\author[2]{Fabrice Jimenez}
\author[1]{Thomas Oberlin}

\affil[1]{Fédération ENAC ISAE-SUPAERO ONERA, Université de Toulouse, Toulouse, France\\
\texttt{\{estelle.chigot2, dennis.wilson, thomas.oberlin\}@isae.fr}}
\affil[2]{Airbus, Toulouse, France\\
\texttt{\{meriem.ghrib, fabrice.jimenez\}@airbus.com}}

\date{} 

\maketitle              
\begingroup
\renewcommand\thefootnote{}\footnotetext{%
\hspace{-2em}
\textit{Accepted in Computer Analysis of Images and Patterns, September 2025 (CAIP 2025).}}
\addtocounter{footnote}{0}
\endgroup

\begin{abstract}
Deep vision models are now mature enough to be integrated in industrial and possibly critical applications such as autonomous navigation. Yet, data collection and labeling to train such models requires too much efforts and costs for a single company or product. This drawback is more significant in critical applications, where training data must include all possible conditions including rare scenarios. In this perspective, generating synthetic images is an appealing solution, since it allows a cheap yet reliable covering of all the conditions and environments, if the impact of the synthetic-to-real distribution shift is mitigated. In this article, we consider the case of runway detection that is a critical part in autonomous landing systems developed by aircraft manufacturers. We propose an image generation approach based on a commercial flight simulator that complements a few annotated real images. By controlling the image generation and the integration of real and synthetic data, we show that standard object detection models can achieve accurate prediction. We also evaluate their robustness with respect to adverse conditions, in our case nighttime images, that were not represented in the real data, and show the interest of using a customized domain adaptation strategy.

\end{abstract}


\section{Introduction}\label{intro}

With the tremendous progress of artificial intelligence, deep vision models are increasingly integrated into critical systems. This paper considers the context of commercial aircraft, where there are major opportunities for both airlines and airports to increase safety and optimize routing or ground operations.
In this work, we consider the task of runway detection from onboard cameras.

Deep learning models require large amounts of data to achieve high performance, but in industrial applications, data are often difficult or expensive to obtain. It may be difficult to reliably collect data, impossible to do so safely, or expensive to gather and label data. Aircraft operations make data gathering even more challenging: data collection may be heavily regulated and dangerous to perform outside of nominal conditions. On the other hand, flight simulators seem to provide the ideal solutions. They are cheap, customizable to include adverse conditions such as night, snow or fog, and, most of all, they can generate corresponding precise labels, thanks to the underlying scene modeling process. However, when using simulated data as the only source of data to train a detection model, a performance drop usually occurs when operating the model in real conditions, since the model cannot generalize to real images. This is known as the domain gap, synthetic-to-real gap, or sim-to-real problem. In this context, we study the reliability and performance gains from using synthetic data in a critical runway detection system.

The problem of runway recognition for aircraft has already been studied in the literature, either as a problem of object detection \cite{wang2022visual,linden2021curating,li2024federated} or semantic segmentation \cite{wang2024valnet}. There are now multiple publicly available datasets for runway detection \cite{ducoffe2023lard,chen2023bars,chen2024image}. However, these datasets usually include few to no real images, limiting the range of applicable methods and the evaluation on real use cases. In this study, we have access to a small private dataset of real images, and a large dataset of synthetic images; in both real and synthetic datasets, we have images representing different environmental conditions. We use these data to explore domain adaptation methods and to test models' robustness to unseen situations. In this work, we focus on runway detection at night as an underrepresented case, as we have access to nighttime images in both datasets, enabling us to evaluate the robustness of runway detection models on real images.

We make the following contributions, within the case of runway detection: (1) we demonstrate the benefits of using synthetic data to train a detection model in an industrial use case; (2) we customize a state-of-the-art domain adaptation method to the runway detection problem; (3) we show the advantages of using a domain adaptation approach to enhance model's robustness under adverse conditions, in our case nighttime.

The paper is structured as follows: \Cref{related-work} reviews the literature in the field of synthetic data for object detection. \Cref{method} details our data collection and the methods compared, while results are presented in \Cref{results}. \Cref{conclusion} concludes the study and draws some perspectives.

\section{Related work}\label{related-work}

\subsection{Object detection}

Object detection is the computer vision task of determining the location and type of objects in a given image.
Recent advances in deep learning have significantly improved object detection, especially in safety-critical industrial applications.
One-stage detectors, such as YOLO \cite{redmon2016you} and SSD \cite{liu2016ssd}, prioritize speed over accuracy by directly predicting object location and class probabilities from the image.
Two-stage models like Faster R-CNN \cite{ren2016faster} perform object detection in two steps. First, an image classification model extracts features from the image; this first model is often referred to as the "backbone." Then, a region proposal network (RPN) generates potential object detection proposals from the features.
Generally, the backbone is pretrained on an image classification task, such as ImageNet \cite{deng2009imagenet}, to improve the model's performance on the object detection task.
In this work, we focus on Faster R-CNN due to its prevalence in the domain adaptation literature and its good performance.

Runway detection is critical for vision-based aircraft landing systems, in order to properly position the aircraft during the procedure. It has been treated as an object detection task in various studies.
Faster R-CNN, in particular, has been used for the detection of runways \cite{linden2021curating}. A fusion of visual and infrared images has been proposed to improve runway detection at night, also using Faster R-CNN \cite{wang2022visual}.
Federated learning, a distributed means of training models, was shown to improve the robustness of runway detection models \cite{li2024federated}. 
One of the key challenges in the application of deep-learning based models for runway detection is the lack of sufficient real-world data, which synthetic data can help address.

\subsection{Synthetic data}
In order to train deep learning models and especially computer vision models, it is necessary to gather a large quantity of data. Large public datasets like COCO \cite{lin2014microsoft} or ImageNet \cite{deng2009imagenet} have accelerated object detection research, but their utility in industrial contexts, such as aviation, is limited. 
However, the cost, safety or legal concerns and time-consuming labeling process of real-world data collection make datasets difficult to build. 
Synthetic data are then a valuable alternative to create more data.

Two main paradigms exist for data generation. The first one, domain randomization \cite{tobin2017domain,tremblay2018training,hinterstoisser2019annotation}, implies that photorealism doesn't matter as it tries to make data as diverse as possible. Simulators or 3D game engines are used to generate objects with large variations of environmental conditions, lighting or textures, improving model robustness across a wide range of scenarios. The aim is to train the model to perceive real images as just another variation of these conditions, leading to strong detection performance. 

The second approach is to generate images as close to the real world as possible. This can be achieved by using computer graphics and simulators to build world replicas, showing realistic physics and real world behaviors. This method is very popular in the automotive industry, where video games like GTA-V \cite{richter2016playing,johnson2016driving} or world models developed in 3D engines \cite{ros2016synthia} have been used to generate widely utilized datasets. Additionally, whole driving simulators \cite{dosovitskiy2017carla,lin2022capturing} have been released for this purpose.

Generative models \cite{goodfellow2014generative,kingma2013auto,sohl2015deep} are also a powerful way to create photorealistic pictures. Generative adversarial networks (GANs) and variational autoencoders learn to approximate the data distribution of their training dataset, while diffusion models are trained to denoise step-by-step random noises into coherent data samples. GANs especially have been leveraged to generate data for industrial applications \cite{niu2020defect,li2022fabric,zhang2022anomaly}. However, generative models need lots of real-world input data to produce satisfactory results. Also, they can't generate labels automatically. In this work we will focus on simulators for those two reasons. 

In the aviation industry, due to the lack of real open data, datasets have been released to facilitate research in this field. LARD \cite{ducoffe2023lard} provides a generator for runway detection, based on Google Earth, as well as 1800 real images taken from YouTube videos and manually labeled. On other tasks than runway detection,  FS2020 \cite{chen2024image} provides a synthetic dataset generated with Microsoft Flight Simulator, meant for runway segmentation and lines detection. Finally, Rareplanes \cite{shermeyer2021rareplanes} integrates 3D models of aircraft into real images for remote sensing applications. Logically, the works published for runway detection use mainly synthetic data. They either create their own with flight simulators \cite{wang2022visual}, or use available open datasets \cite{li2024federated} such as LARD. In a try to assess the benefits of synthetic images, Linden et al. \cite{linden2021curating} compare the impact on detection performance of several environmental conditions in a synthetic dataset and examines a style transfer approach on automotive data.

\subsection{Synthetic to real domain adaptation}

Domain adaptation (DA) methods attempt to minimize the synthetic to real gap by bringing closer either the images directly or the feature distributions between domains. Following the first option, SC-UDA \cite{yu2022sc} uses a neural style transfer method to apply a realistic style on synthetic images, and then uses Faster R-CNN to generate bounding boxes to augment the unlabeled real dataset. On the other hand, CARE \cite{prabhu2023bridging} tries to align the instances features between synthetic and real images by modifying the traditional Faster R-CNN loss function. Using the available labels, the loss term introduced by CARE computes the difference between the features of a synthetic object and a real one, and tries to minimize it along the other object detection loss terms. CARE also introduces two reweighting terms to adapt to the difference of object sizes and frequency between the two domains.

In the context of runway detection, DA remains underexplored, likely due to the scarcity of real-world data. In this work we focus on studying DA methods to enhance model generalization from synthetic to real images.

\section{Methodology}\label{method}

\subsection{Data collection}
In our use case, real images are limited in quantity due to their high cost of acquisition. The images are captured using an aircraft equipped with cameras and sensors, navigating across various airports in the U.S. Data is acquired by photographing runways at selected airports during landing sequences. In this study, we have access to a database of real images from 27 airports with their associated metadata. However, the images of the same airports tend to be quite similar, showing the same runways from close points during the flight phases, resulting in limited diversity within the real data. Labeling is performed automatically based on the GPS positions of the aircraft and the runways, leading to potential errors in labels due to factors like miscalibration or cloud occlusions.

In contrast, we generate our own synthetic dataset using the commercial flight simulator XPlane12, which allows us to specify the aircraft's position, orientation, and various environmental conditions such as weather and time. This customization capacity enables us to collect diverse images from 207 airports worldwide. One of the key benefits of synthetic images is the ability to achieve automatic and accurate labeling of bounding boxes and other parameters that typically require human input in real-world scenarios. We use a custom pipeline rather than relying on existing datasets because it enables the generation of a large volume (>5,000) of high-quality images, across a wide variety of airports and under specific environmental conditions.

Taking advantage of this generation capacity, we want to trial synthetic data and assess if using simulated images enables a model to be robust under unseen conditions. To do so, we use the nighttime scenario. At night, runways are very different than during the day. The light conditions differ a lot as they are outlined by lights, in order to enable the pilot to see the runway. The shape is conserved but the texture and colors of the image are not, which is a good way to assess the generalization capabilities of a model. With this idea in mind, we generate nighttime synthetic data from the same airports as the daytime ones.

\begin{figure}
\centering
\begin{subcaptionblock}[c]{0.45\columnwidth}
    \includegraphics[alt={real runway taken from an aircraft}, width=0.9\linewidth, trim= 2cm 12cm 0cm 1cm, clip]{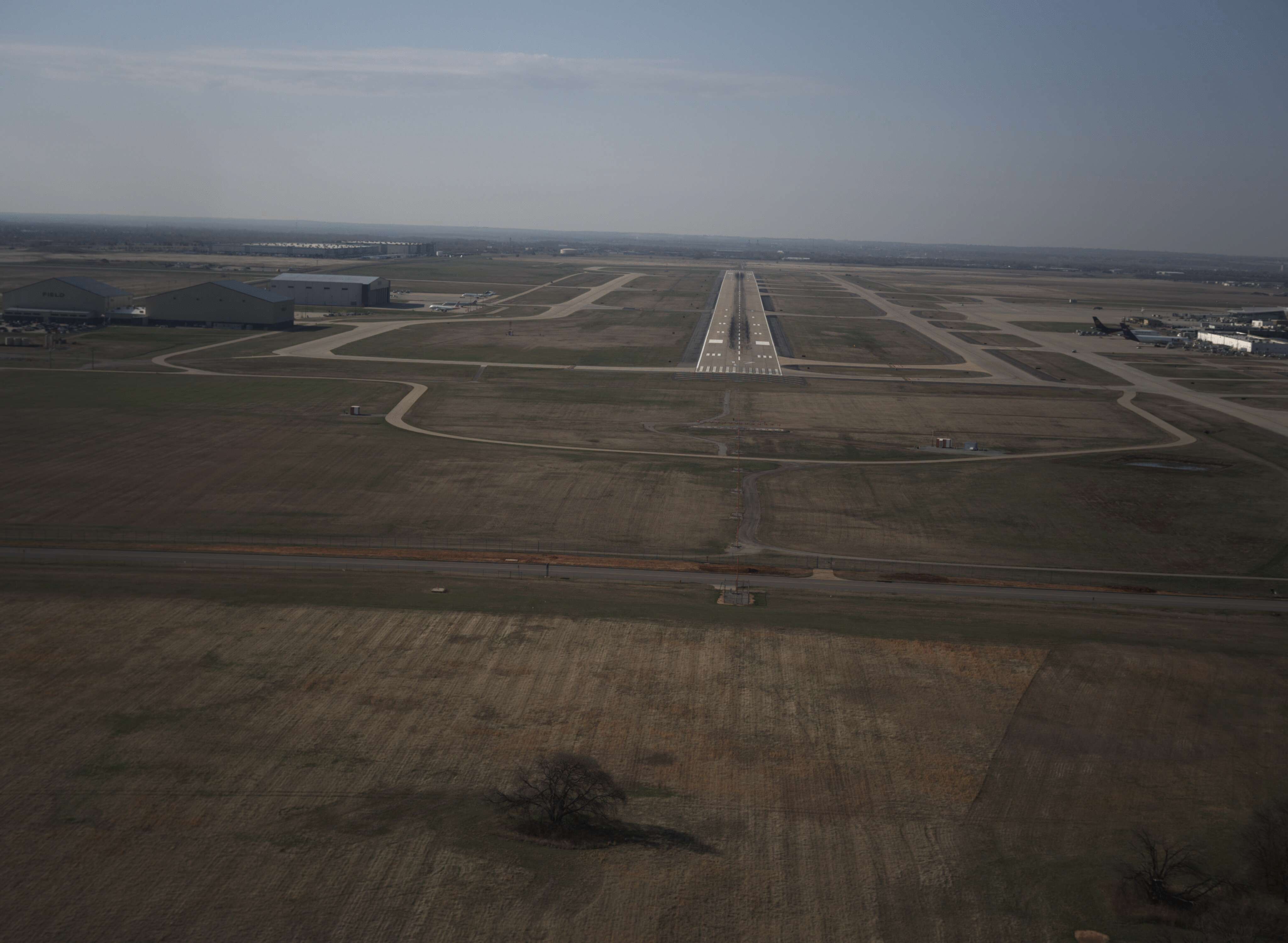}
    \centering
    \caption{Real runway (Source: Airbus)}
\end{subcaptionblock}
\begin{subcaptionblock}[c]{0.45\columnwidth}
    \includegraphics[alt={synthetic runway from a simulator}, width=0.9\linewidth, trim= 10cm 4cm 10cm 4cm, clip]{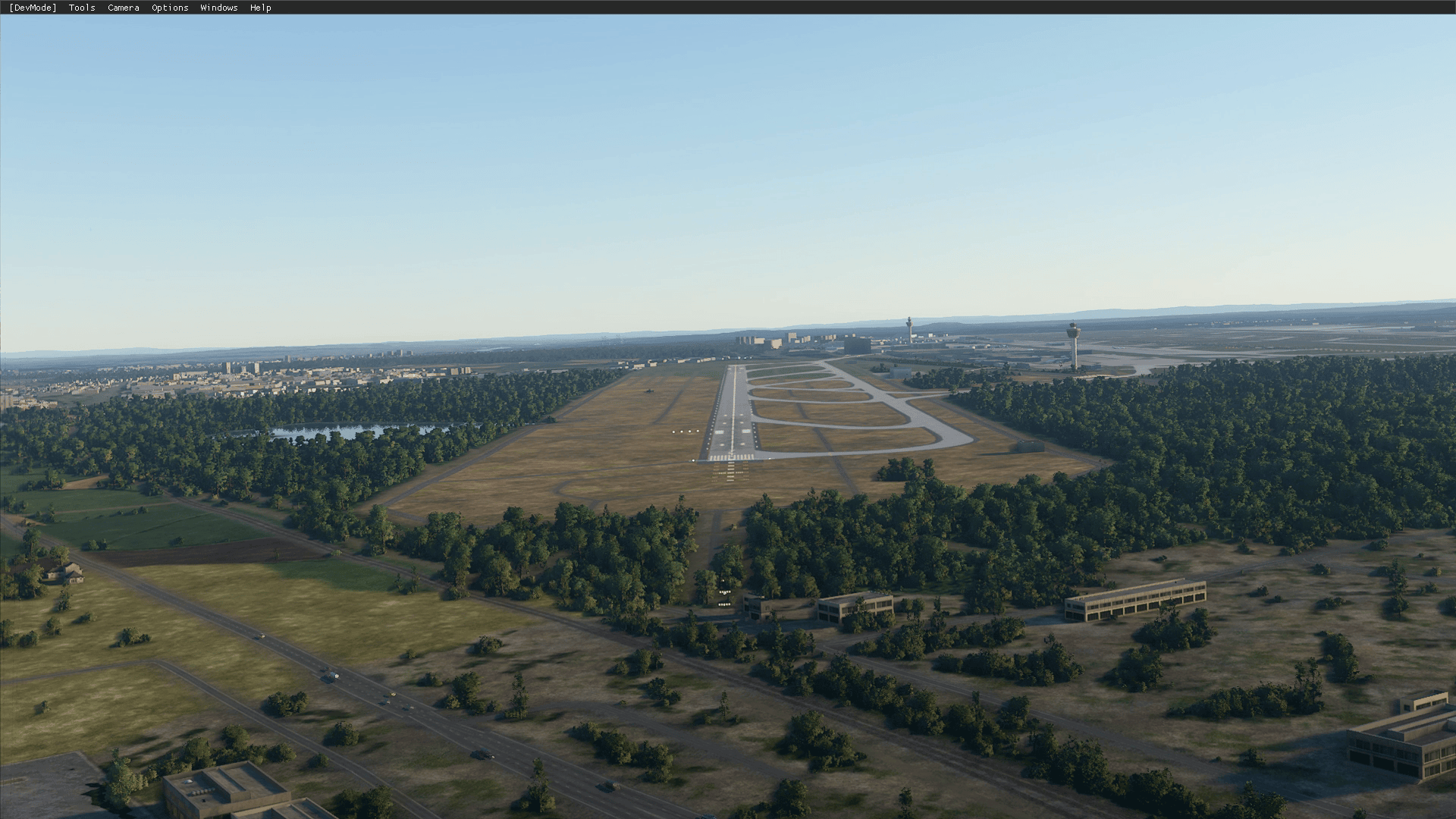}
    \centering
    \caption{Synthetic runway (Source: \cite{chen2024image})}
\end{subcaptionblock}
\begin{subcaptionblock}[c]{0.6\columnwidth}
    \includegraphics[alt={synthetic runway at night from a simulator},width=0.6\linewidth, trim= 10cm 4cm 10cm 4cm, clip]{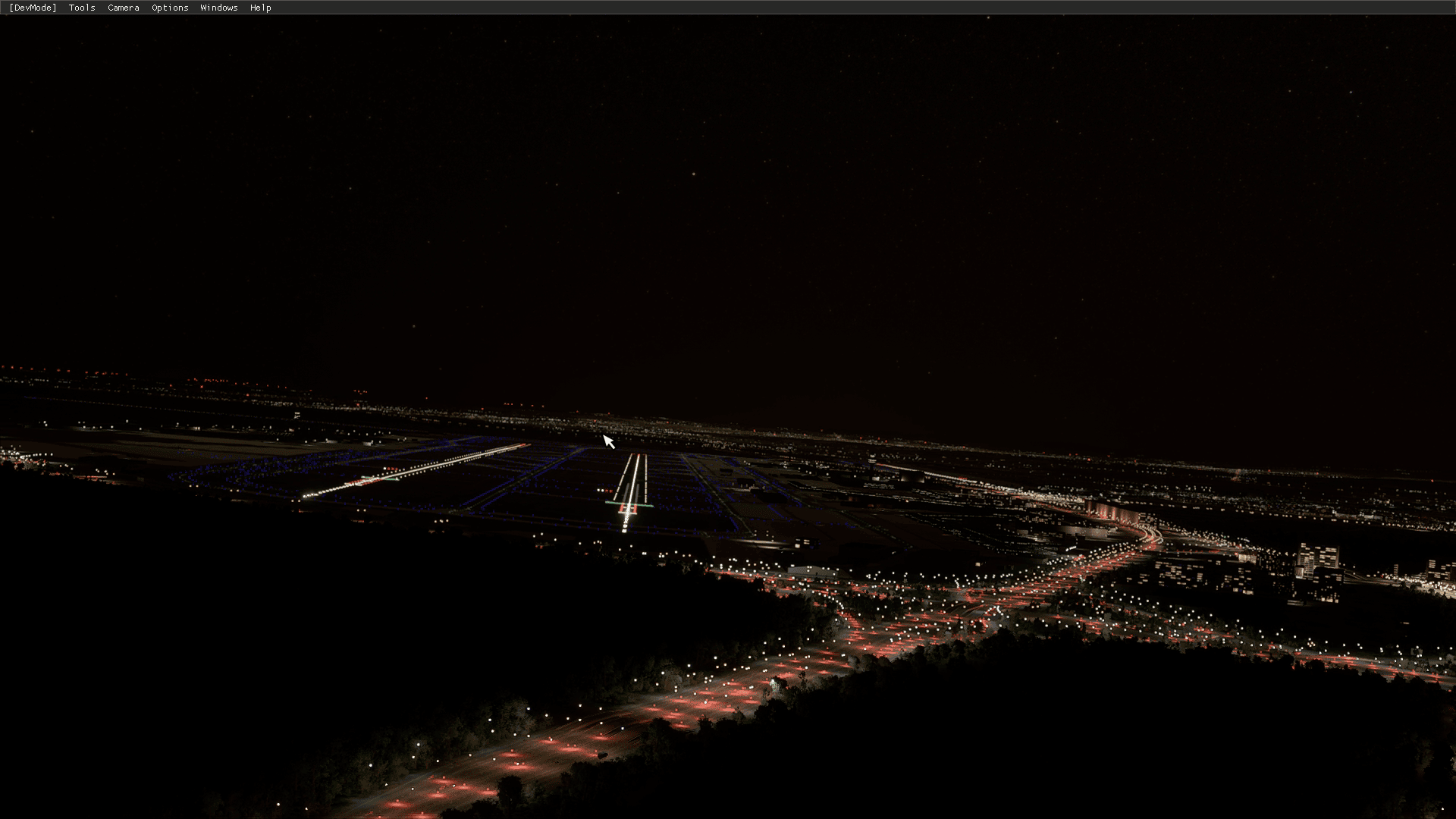}
    \centering
    \caption{Synthetic runway at night (Source: \cite{chen2024image})}
\end{subcaptionblock}
\caption{Examples of real data and synthetic data for runway detection.}
\label{image_examples}
\end{figure}

In \Cref{image_examples} we have examples of real and synthetic images for the runway detection task. For copyright reasons, we show synthetic images from \cite{chen2024image} which are not the one used to train the model in this study, but closely resemble. 
\subsection{Mixing strategies}
In this work, our objective is to assess the use of synthetic data for runway detection. We also want to study the impact of synthetic data in unseen situations. To this end we design several training strategies, applicable for both the nominal and the nighttime studies.

We explore the utility of synthetic data by employing various configurations of real and synthetic datasets. Our study begins with a baseline experiment using only real images of runways (\mreal), allowing us to establish a performance benchmark based solely on real data.

Next, we create a synthetic dataset using a flight simulator software (\msynth). For all datasets, we ensure that each airport is represented by the same number of images, preventing imbalances in both training and validation sets.

To integrate both real and synthetic data during training, we assess different mixing strategies. The standard approach involves randomly sampling from the combined dataset (\mmix), resulting in a variable ratio of synthetic to real data at each training step, dependent on the dataset composition.

To mitigate the domain shift between synthetic and real data, we also examine the CARE \cite{prabhu2023bridging} domain adaptation method (\mcare) tailored to our use case. This method introduces an additional loss term during training, encouraging the detection model to represent real and synthetic objects similarly by minimizing the Euclidean distance between their feature maps.

Furthermore, the CARE framework employs a specific sampling strategy that ensures an equal number of synthetic and real images are used at each training step in the minibatch. While this sampling method is necessary to their loss terms, we also investigate its effectiveness independently, without the accompanying loss terms (\msampler).

\subsection{Experimental setup}

\paragraph{Model overview}
We consider a standard object detector, Faster R-CNN \cite{ren2016faster}, to follow the recent literature in domain adaptation for object detection \cite{prabhu2023bridging,yu2022sc} and previous works in airport runway detection \cite{wang2022visual,linden2021curating,li2024federated}.

The objective function used in this study, which the model tries to minimize, is the standard Faster R-CNN loss function, $\ell_{F-RCNN}$. 



In the experiments using \mcare, we add to the standard loss the alignment term proposed. \mcare\ also incorporates reweighting factors that are unnecessary in our monoclass setting where object size and aspect ratio are consistent. We do not include them in our loss function. The overall training loss then becomes:
\begin{equation}\label{care_eq}
    \ell_{det} = \ell_{F-RCNN} + \lambda\ell_{align}.
\end{equation}


$\lambda$ is a balancing factor, which controls the contribution of the alignment loss in the overall loss. $\ell_{align}$ is the alignment loss proposed by \mcare, a cycle loss computing the difference between features of synthetic and real objects. 

\paragraph{Implementation details}
For all the experiments using the real dataset, our data being redundant, we use a small training dataset of 1,000 images from 19 airports. Then, for the simulated dataset we generate 10,000 images to keep a reasonable ratio of synthetic to real images. Those images are from 199 airports, including the ones in the real dataset.

In order to study the robustness to domain shift of each model, we introduce the nighttime condition. For this purpose, we generate another synthetic dataset composed of 5,000 daytime images and 5,000 nighttime images from the same 199 airports.

Our detection model is a Faster R-CNN with FPN and a ResNet50 backbone, pretrained on COCO2017 \cite{lin2014microsoft}. The training configuration includes a batch size of 8, a learning rate of 0.002, and is trained during 10,000 iterations. We use $\lambda = 0.1$ for the CARE loss (\Cref{care_eq}). We perform the validation at the end of the training.

\paragraph{Evaluation}
Following previous work \cite{yu2022sc,linden2021curating} we report COCO Average Precision (AP) for each strategy, AP being the average precision over several Intersection over Union values, from 0.5 to 0.95 by 0.05 steps.

To assess the training methods and the impact of training datasets, we build an evaluation dataset of 200 real daytime images from 8 airports. Those airports are not represented in any training set. In order to evaluate robustness to the nighttime condition, we build an additional validation dataset of 200 real nighttime images from the same 8 airports. Real nighttime images are never seen during training.

\section{Results}\label{results}
\subsection{Comparison of training strategies}

\begin{table}
\centering
\begin{tabular}{| c | c | c | c | c | c |}
\hline
 & \mreal & \msynth & \mmix & \msampler & \mcare \\
\hline
DAY AP & 58.60 & 59.04 & \textbf{65.25} & 64.01 & 63.66 \\
\hline
\end{tabular}
\caption{ 
Values of AP for each mixing strategy over daytime validation datasets. Models were trained \textit{without} synthetic nighttime image.}
\label{fig_day_val_day}
\end{table}
In this section, we evaluate models trained on daytime images only. 
\Cref{fig_day_val_day} compares the results of the different training strategies and datasets on the daytime validation set.
We can observe that all methods outperform the model trained on real data, including the synthetic only experiment (+0,44\%). We could explain this behavior with the model's pretraining; having already seen real images in COCO2017, it acquired features relevant to real images before the synthetic training.
In this setting, adding synthetic data into the training dataset seems to bring better performance on the validation dataset, and a simple method such as \mmix\ seems to bring the best results (+6.65\% compared to \mreal\ only). This conclusion was expected, as bringing more data with reliable labeling should lead to a better model.

\subsection{Results on night domain shift}
We analyze the results of models integrating nighttime images in their synthetic training set, compared to the same models trained on daytime images only.

\begin{table}
\centering
\begin{tabular}{|c|c|c|c|c|c|}
\hline
 & \mreal & \msynth & \mmix & \msampler & \mcare \\
\hline
DAY AP & 58.60 & 57.55 & 66.77 & 64.73 & \textbf{67.50} \\
\hline
NIGHT AP & 15.21 & 43.22 & 43.38 & 42.00 & \textbf{43.75} \\
\hline
\end{tabular}
\caption{Values of AP for each mixing strategy over daytime and nighttime validation datasets. Models were trained \textit{with} synthetic nighttime image.}
\label{fig_daynight_val_day}
\end{table}

In \Cref{fig_daynight_val_day}, we first compare the models performance on the daytime validation dataset. The \mreal\ model keeps the same training set throughout all experiments and therefore gets the same results. The synthetic only model \msynth\ suffers from the addition of nighttime condition. On the contrary, while we expected the same results from the other training strategies, but \mmix\ (+8,17\%) and especially \mcare\ (+8,90\%) seem to benefit from the diversification of conditions. On average \msampler\ attains the same results, showing little use of the nighttime images. Overall all mixing methods still outperform the \mreal\ experiment. The \mcare\ method exhibits the best score showing the benefits of this domain adaptation strategy when diversifying the training dataset.

We also evaluate our models over the nighttime validation dataset. The \mreal\ model experiences a significant drop in performance (-43,39\%), and is unable to generalize properly on this unseen condition. All the other methods also experience a drop in performance. However, they are all able to maintain an AP score similar to the \msynth\ model (-14,33\% compared to daytime validation), way higher than the \mreal\ model on both validation set. In this scenario, adding synthetic data seems to guarantee a correct level of performance, compared to a completely unseen condition. Our customized \mcare\ method still exhibits the best score in this difficult scenario by a small margin. 

\begin{figure}
\centering
\captionsetup{subrefformat=parens}
\begin{subcaptionblock}[c]{0.25\columnwidth}
    \centering
    \includegraphics[alt={real runway at night, ground truth}, width=\columnwidth, trim= 5cm 10cm 8cm 8cm, clip]{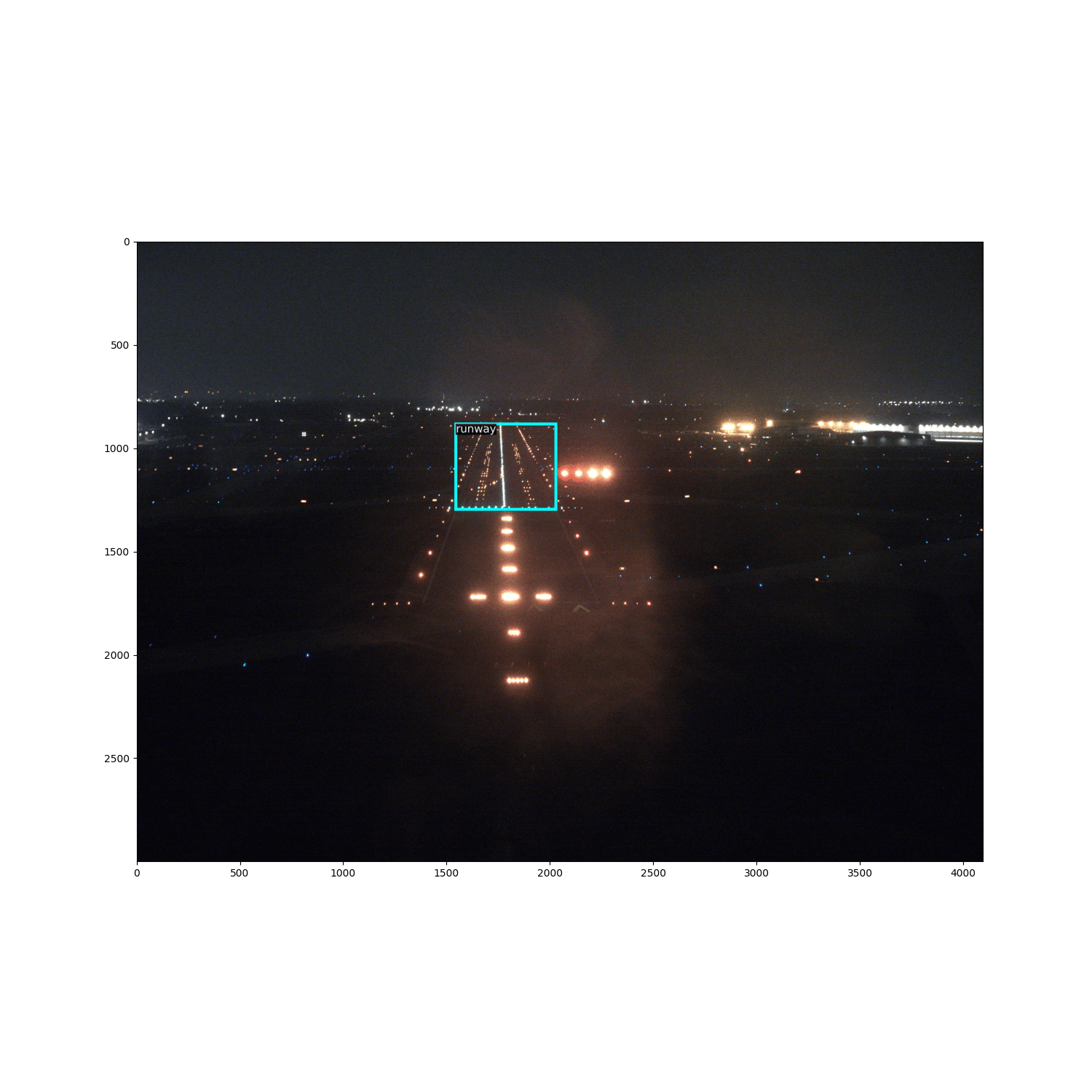}
    \caption{Ground truth}
    \label{ground_truth}
\end{subcaptionblock}
\begin{subcaptionblock}[c]{0.25\columnwidth}
    \centering
    \includegraphics[alt={real runway at night, real model inference}, width=\columnwidth, trim= 5cm 10cm 8cm 8cm, clip]{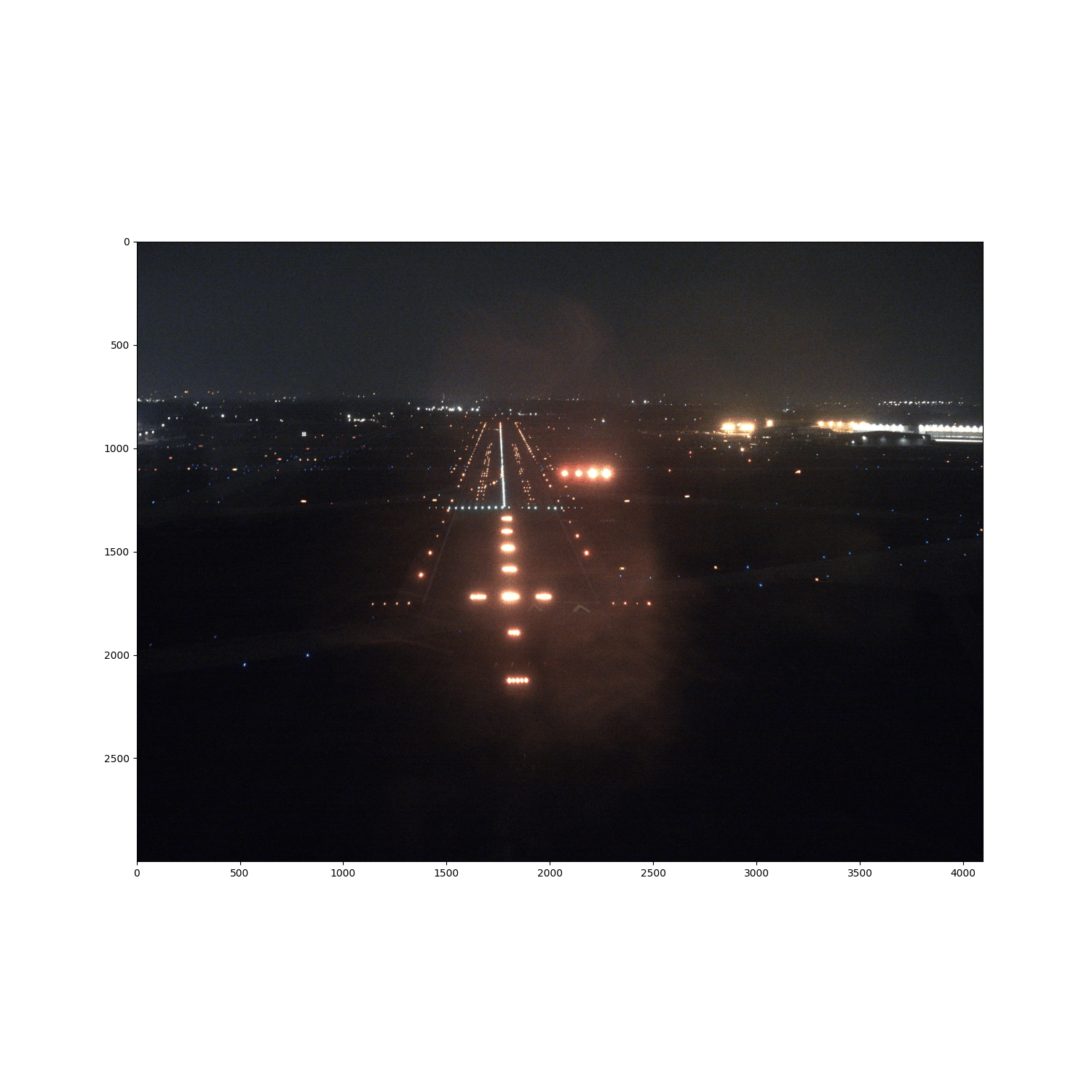}
    \caption{\mreal}
    \label{inference_real}
\end{subcaptionblock}
\begin{subcaptionblock}[c]{0.25\columnwidth}
    \centering
    \includegraphics[alt={real runway at night, synth model inference}, width=\columnwidth, trim= 5cm 10cm 8cm 8cm, clip]{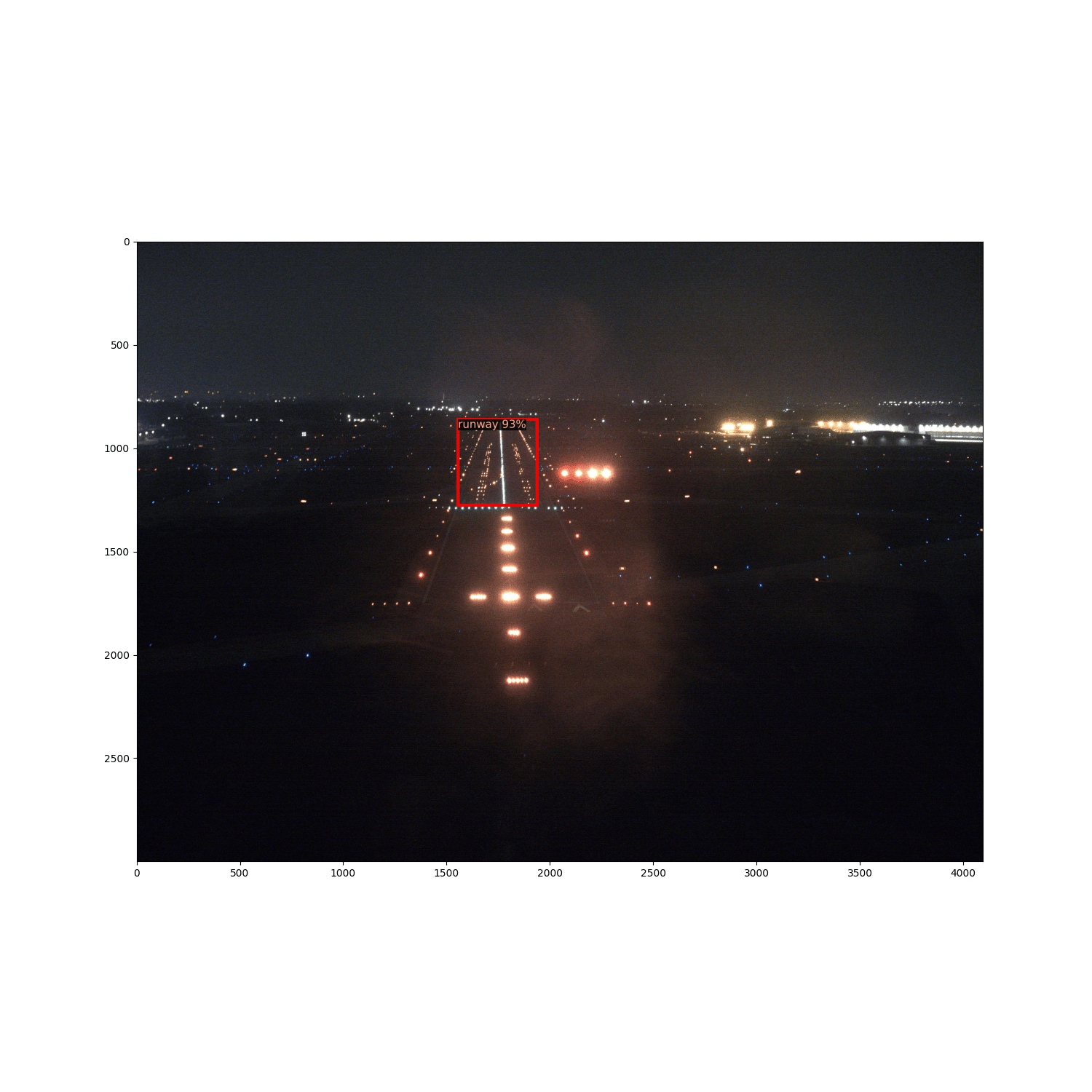}
    \caption{\msynth}
    \label{inference_synth}
\end{subcaptionblock}\\
\begin{subcaptionblock}[c]{0.25\columnwidth}
    \centering
    \includegraphics[alt={real runway at night, mix model inference}, width=\columnwidth, trim= 5cm 10cm 8cm 8cm, clip]{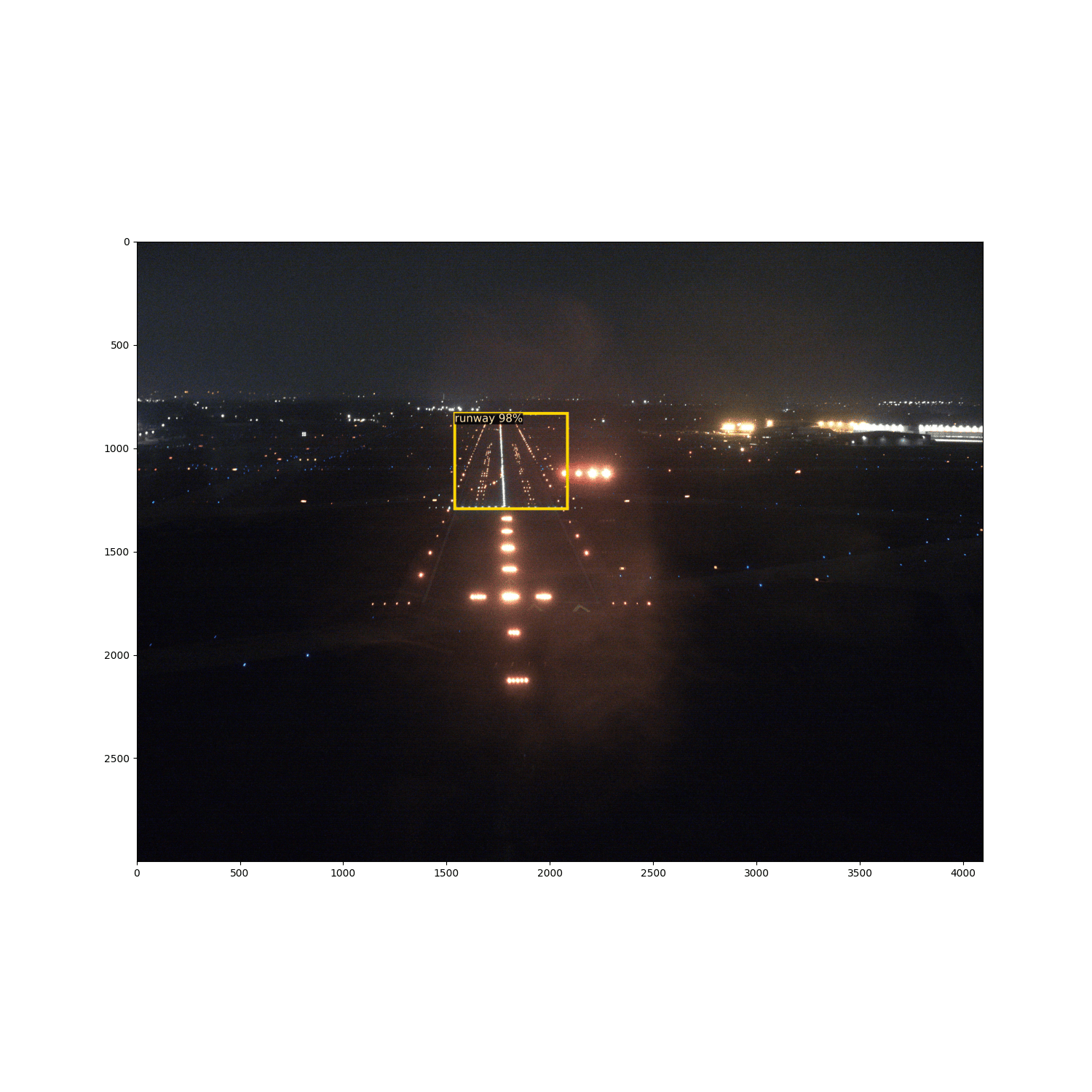}
    \caption{\mmix}
    \label{inference_mix}
\end{subcaptionblock}
\begin{subcaptionblock}[c]{0.25\columnwidth}
    \centering
    \includegraphics[alt={real runway at night, sampler model inference}, width=\columnwidth, trim= 5cm 10cm 8cm 8cm, clip]{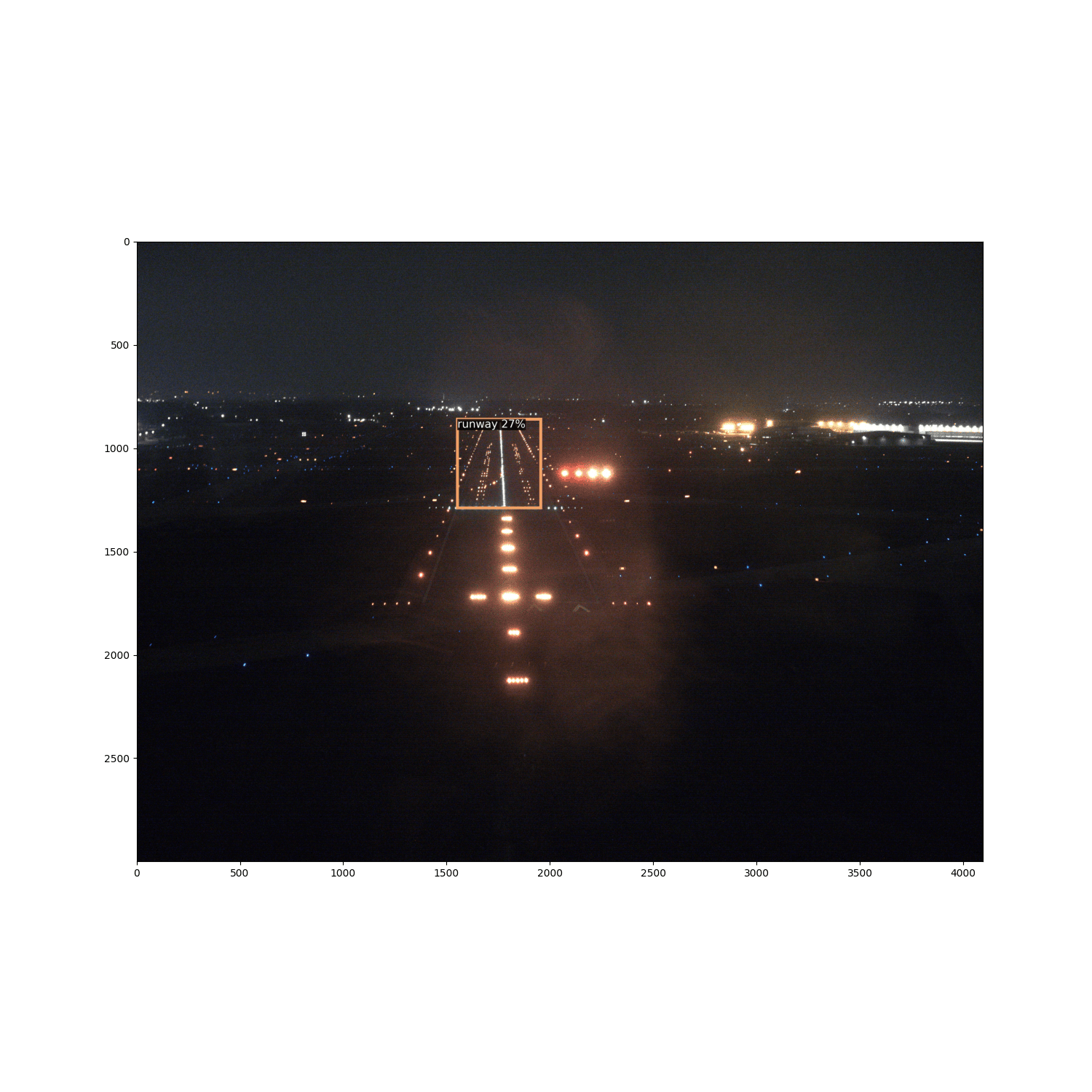}
    \caption{\msampler}
    \label{inference_sample}
\end{subcaptionblock}
\begin{subcaptionblock}[c]{0.25\columnwidth}
    \centering
    \includegraphics[alt={real runway at night, care model inference}, width=\columnwidth, trim= 5cm 10cm 8cm 8cm, clip]{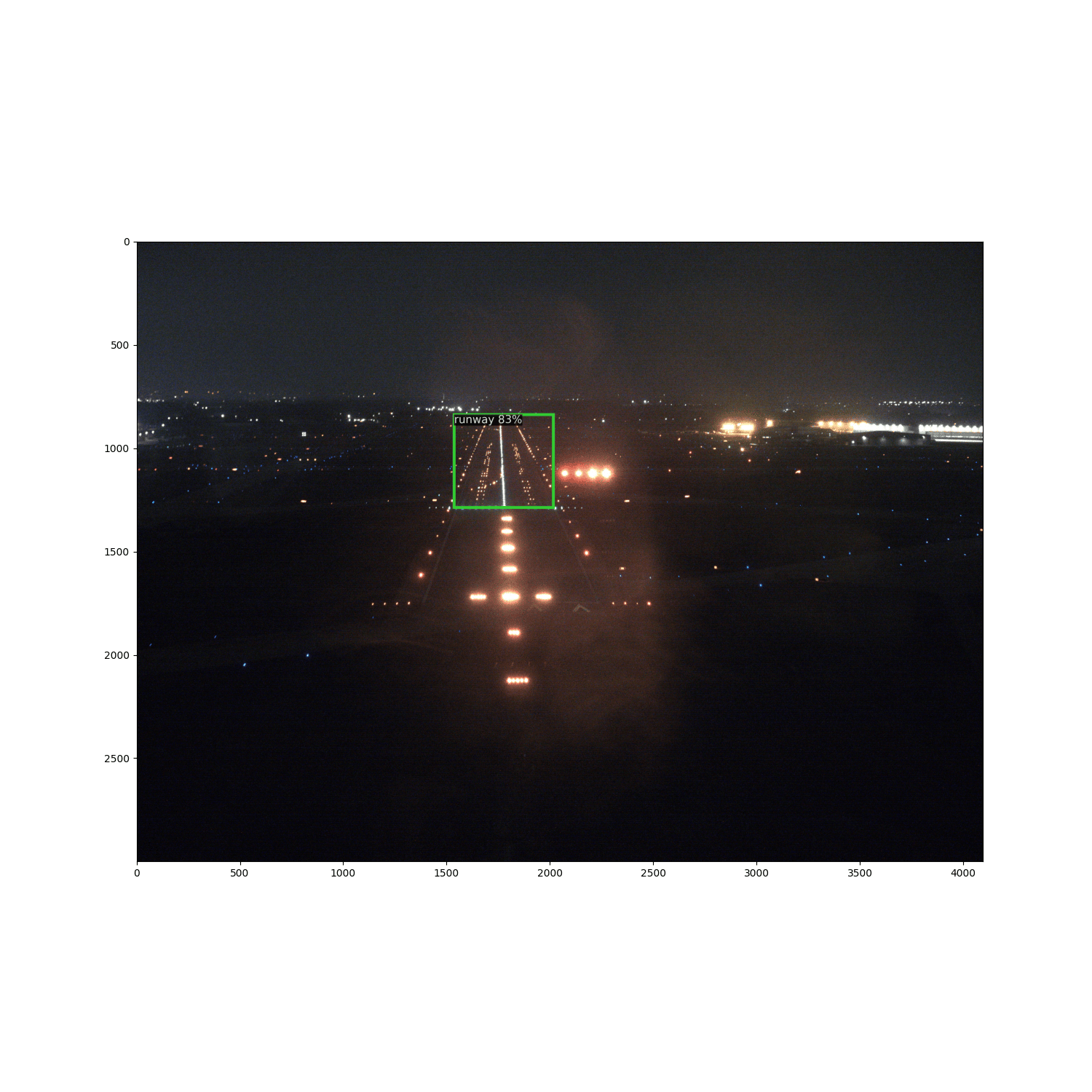}
    \caption{\mcare}
    \label{inference_care}
\end{subcaptionblock}
\caption{A real image of an airport runway at night from the validation dataset, and the bounding box determined by inference through each model. The \mreal\ model does not predict any bounding box, while all other models correctly predict the bounding box covering the runway.}
\label{inferences}
\end{figure}

In \Cref{inferences} we show inference results for all models on one validation image. This picture example is representative of the type of nighttime images we can find in the dataset. The \mreal\ model is unable to see the runway whereas all the models trained on synthetic data could detect it.

Overall, mixing synthetic and real data seems to lead to more robustness to unseen environmental conditions, as we obtain good AP scores in both nominal and adverse conditions. Adding the target data into the synthetic datasets helps getting reasonable performance in difficult conditions.
In our case, domain adaptation seems to improve performance with the customized loss of \mcare\ allowing the model to generalize better. While the sampling method introduced in \mcare\ seems to lower the detection performance, its alignment loss compensates this drop and improves the final results when getting enough diversity in the training dataset.

\section{Conclusion and perspectives}\label{conclusion}

In this work, we studied the use of synthetic data for runway detection, specifically how data sampling schemes and domain adaptation can improve model performance. We find that, in general, synthetic data improves model performance and robustness in adapting to rare conditions.

In our first study on the nominal case of regular daytime conditions
we find that synthetic images can improve model accuracy, but the presence of both real and synthetic images is necessary. 

Synthetic data can be especially valuable to represent rare cases that are underrepresented in real data, in order to ensure model robustness. 
We found that synthetic images of nighttime conditions can greatly improve model performance on runway detection at night. 
Even training a model on only synthetic data vastly outperforms models trained only on real data.
We consider that this is due to the lack of real nighttime images and the difficulty of adapting runway detection from daytime to nighttime in the absence of nighttime examples.

Domain adaptation seems to be a promising approach to enhance runway detection models. While the benefits of \mcare\ compared to a simple mixing approach seem modest, the customized loss helps overcoming the decline introduced by the sampler, and \mcare\ achieves the best performance overall.

In a future direction we aim to tackle other adverse conditions, such as snowy or foggy weathers that are more complicated to capture in reality or render in simulation. It would also be of interest to explore the impact of the \mcare\ sampler, to further improve the benefits of this domain adaptation approach.

In summary, we show that synthetic data can aid in the task of airport runway recognition. Images from simulation software are especially useful in the representation of rare cases, such as nighttime data. By including synthetic data from diverse cases, we can ensure and improve computer vision model robustness, especially with the help of our customized \mcare\ domain adaptation strategy.


\paragraph{Disclosure of Interests}
Estelle Chigot, Meriem Ghrib and Fabrice Jimenez are employees of the Airbus company. Dennis G. Wilson and Thomas Oberlin have no competing interests to declare that are relevant to the content of this article.

\paragraph{Acknowledgement}
This preprint has not undergone peer review or any post-submission improvements or corrections. The Version of Record of this contribution is published in Lecture Notes in Computer Science, and is available online at \url{https://doi.org/10.1007/978-3-032-04968-1_25}.

%
%
%
\bibliographystyle{splncs04}
\bibliography{biblio}
\end{document}